\documentclass[runningheads]{llncs}
\usepackage{textcomp}
\usepackage{graphics} 
\usepackage{times}
\usepackage{amssymb}  
\usepackage{graphicx}
\usepackage{color}
\usepackage{subfigure}
\usepackage[ruled]{algorithm2e}
\usepackage{cite}
\usepackage{makecell}
\usepackage{subfigure}
\usepackage[utf8]{inputenc}
\usepackage{epigraph}
\usepackage[width=122mm,left=12mm,paperwidth=146mm,height=193mm,top=12mm,paperheight=217mm]{geometry}
\usepackage[colorlinks,allcolors=black,pdftex]{hyperref}

\begin{document}
\pagestyle{headings}
\mainmatter
\def\ECCV18SubNumber{3009}

\title{Sorted Pooling in Convolutional Networks for One-shot Learning}






\titlerunning{Sorted Pooling in Convolutional Networks for One-shot Learning}

\authorrunning{Horváth András}

\author{Horváth András}
\institute{Peter Pazmany Catholic University}

\maketitle

\epigraph{"This idea that there is generality in the specific is of far-reaching importance."}{\textup{ Douglas R. Hofstadter}: Gödel, Escher, Bach}

\begin{abstract}
We present generalized versions of the commonly used maximum pooling operation: $k$th maximum and sorted pooling operations which selects the $k$th largest response in each pooling region, selecting locally consistent features of the input images. This method is able to increase the generalization power of a network and can be used to decrease training time and error rate of networks and it can significantly improve accuracy in case of training scenarios where the amount of available data is limited, like one-shot learning scenarios.
\end{abstract}

\section{Introduction}

The renaissance of neural networks was and currently still is boosted by convolutional neural networks \cite{krizhevsky2012imagenet}. These networks caused significant breakthroughs in many practical problems which were not solvable before. These methods require a large and general dataset (representing the distribution of the real problem) to select kernels which provide the highest accuracy on our training- and test-set. In many practical problems only limited amount of data is available or limited amount of data can be presented to the network during training in one step, because of memory constraints. Learning scenarios in case of limited data can be helped by data augmentation or by using special architectures (matching networks\cite{vinyals2016matching}, memory augmented neural networks\cite{santoro2016one}) to solve these problems.
Even in case of a sufficiently large dataset bias-variance problem \cite{geman2008neural} is one of the major and most significant challenges of machine learning. We can never be sure that the data we use is general and matches exactly the real world distribution of possible input images of the problem. 
We want to achieve two goals simultaneously with every training process: we would like to achieve the highest possible accuracy on our training, test and evaluation set, but in practice we also want to create a general method. Because of this, we want to select features which ensure high accuracy on the dataset, but also general features which are not "too specific" to our data. Our final aim is always the creation of a general model which works well in practice which after a point is not always the same as having the highest accuracy on our selected train and test sets.
In this paper We present a method which can help creating balance between the selection of features providing high accuracy and choice of general features.

\subsection{Building blocks of convolutional neural networks}

The three main building blocks of Convolutional Neural Networks (CNN) are the following: Convolution, Non-linearity and Pooling.
We have to note that there are many more commonly applied elements and layers like batch normalization \cite{ioffe2015batch} or dropout \cite{srivastava2014dropout} which can be used to increase the generalization capabilities and robustness of a system or other structures like residual networks \cite{he2016deep} or all convolutional networks \cite{springenberg2014striving}, but these three elements are applied in almost every commonly used networks.

Common networks are built by creating a chain (in more general case an acyclic graph) of these three operations. Not only the graph of these elements -the structure of the network- can alter the complexity of the function which the network can approximate, but also these elements may vary with respect to the applications. This paper focuses on the pooling operation and provides a generalized pooling method which performs better in one-shot learning scenarios than maximum or average pooling.

The first neural networks applied average pooling, where a feature map was down-sampled and each region  was substituted by the average value in the region:
\begin{equation}
   P_{avg}(I_{i,j})=\frac{1}{N} \sum_{k,l \in R_{i,j}} I_{k,l}
\end{equation}
Where $P_{avg}$ is the pooling operator. $I$ is the input feature map, $R$ is a two-dimensional region which is selected for pooling, $N$ is the number of elements in the region. 
The notation uses two-dimensional pooling and feature maps, because the operation is usually applied on images and two-dimensional inputs, but can also be used with one- or higher-dimensional data.
This notation focuses only on the pooled region and does not deal with the stride of the convolution operation, which can be used to set the overlapping area between pooling regions, but we consider this as a hyper-parameter of the network architecture and not an inherent part of the operator.
Average pooling considers the whole input region and all values are used to create the response of the pooling layer which was beneficial in the selection of general features. On the other hand Average pooling is a linear operator and can be considered a simple and special case of a convolution kernel and two subsequent convolutions can be replaced by one larger convolution kernel.  Later it was substituted in almost every application by maximum pooling where the maximum is selected from each region. 
\begin{equation}
    P_{max}(I_{i,j})= max(R_{i,j})
\end{equation}
Maximum pooling performs well in practice, adds extra non-linearity to the network and is easy to be calculated (only an index has to be stored to propagate the error back).
Maximum pooling also has disadvantages. It results a really sparse update in the network. Only the neuron resulting the maximum element in the kernel will be responsible for the update of the variables during backpropagation and the activation of other neurons does not matter at all. There is a tendency in certain networks like  variational autoencoders \cite{kingma2013auto}, generative adversarial networks (GANs)\cite{goodfellow2014generative}, or networks used for one shot learning \cite{fei2006one} to avoid pooling because it grasps only certain elements of the input data which results features which are not general enough.
Modified pooling methods has also appeared in segmentation problems like ROI pooling in \cite{ren2015faster} or \cite{he2017mask} which helps in the more accurate localization of regions of interest but does not help in the selection of features inside the proposed regions, where usually maximum pooling is used.
An other approach is introduced in \cite{murray2014generalized} which uses patch similarity over batches to generate a combination of weighted and maximum pooling, but this method also requires large batches and a large amount of data and can not be applied in one-shot learning scenarios. \cite{lee2016generalizing} also proposes the application of heterogeneous pooling methods  inside the network (e.g.: average pooling in certain layers and max pooling in other) but the selection of pooling methods for each layer is a hyper-parameter of the network and is difficult to be optimized in practice.
Here We propose a generalized pooling-method which can help keeping more activations and by this detecting more general features in every region.

\section{$k$th Maximum-Pooling and Feature Consistency}\label{SortedPooling}

The main concept which leads to the application of $k$th maximum-pooling is that convolutional networks are based on local information and exploit the local relations of the data.
Natural images are locally consistent. If a convolutional kernel gives a high response in a region, one can expect that the kernel will give similar results around this region. Small perturbations can not change the response of the network abruptly. 
That is the original and main concept of pooling \cite{lecun1995convolutional} - to balance small variances caused by translation, but doing maximum-pooling neglects all other activations and selects only a single element with the highest response. None of the later features are important, the network can easily be overtrained, because it uses a single example from the image in the pooled kernel. In case of a large amount of data these differences and noises are averaged out, because the noises are usually different on all images, and the selection of general features comes from this average. Those features will be selected which can be found on most of the images. Unfortunately in case of one-shot learning scenarios with limited data, these general features are difficult to be found and has to be estimated for a smaller sample.

Natural images consist of alternating smaller and larger homogeneous regions. CNNs detect the changes between these regions, but even areas at alternations, edges/gradients between these regions contain structure. The appearance of similar structures and periodicity is also a characteristic feature of natural images. This is a reason why natural images in most cases can be reproduced fairly well form low-frequency Fourier components and follow log distribution in the Fourier domain \cite{hou2007saliency}. This is a difference between natural images and general noise. In most cases we want to classify and detect these structural elements on images.
In both smaller and larger regions the responses of convolutional kernels are usually consistent on trained networks and the appearance of certain features and patterns is repeated. Based on this one can easily see that if a convolution kernel detects general patterns and not caused by noise it will probably appear multiple times in a region.
This can lead us to the expectation that if one would like to detect a general feature in an area it should appear more than once in a region. This assumption goes completely against maximum pooling, which selects only the largest response of a kernel in a region and completely neglects all the other responses.

\begin{figure}[h!tb]
\centering{
\subfigure{\includegraphics[width=3.0in]{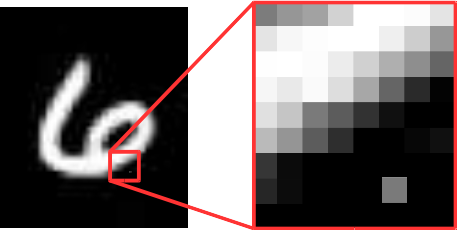}}\\
\subfigure{\includegraphics[width=3.0in]{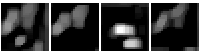}}\\
\caption{ Natural features are consistent. We do not want to select a feature which fits a position perfectly and results a high match. Such features could be too specific for our input image and result overfitting. In the first row an image taken from the MNIST dataset can be seen. We want to select a feature for the detection of diagonal gradient in the depicted region, because it is consistent on the image. Other features, like the small individual dot (appearing in the lower right corner of the image) or a corner-point (appearing in the upper left corner of the image) might give high response but are not consistent in the region. It could easily happened that these patterns are first detected by randomly initialized kernels and would be optimized in maximum-pooling, but these errors could be detected if we would require that a kernels has to give more than one high responses in a region. In the second row four activation maps can bee seen in the third layer of a CNN from an MNIST data. As it can be seen regions are activated in this deeper layer as well, instead of the activation of individual neurons.)
}\label{FeatureConsistency}
}
\end{figure}

It is also worthwhile to note that appearance of locally consist kernel responses can also be observed in deeper layers. Although neurons in deeper layers have larger receptive fields and the size of the input data is decreased layer-by-layer during the training of a network, one can usually observe patches and regions of activations in deeper layers as well, instead of the activation of individual neurons. (Of course this can be changed by changing the size and stride of pooling kernels, but in case of common architectures the activation of regions can be observed  even in deeper layers.) A simple presentation of this fact can be seen by visualizing the activation in deeper layers. An example image depicting feature consistency and activations can be seen on Fig \ref{FeatureConsistency}, online demonstration of the appearance of these consistent local features can also be seen on the online demos of Andrej Karpathy\footnote{one can examine activations of deeper layers without the installation of deep learning frameworks for the MNIST: http://cs.stanford.edu/people/karpathy/convnetjs/demo/mnist.html and CIFAR dataset: https://cs.stanford.edu/people/karpathy/convnetjs/demo/cifar10.html datasets.}.
The presence of features depend not only on the size and stride of pooling convolution kernels but  also from the resolution and object scale on the input image. In case of practical problems and generally used architectures like Alexnet \cite{krizhevsky2012imagenet}, VGG\cite{simonyan2014very} or Resnet50\cite{he2016deep} even deeper layers contain consistent features, since the input images usually contain objects in various scales and our aim is to detect the objects both in small and larger scales.

Instead of selecting local features we introduce  $k$th maximum-pooling which selects the $k$th largest response in every region.
$k$th maximum-pooling can significantly increase the generalization power of a network in case of limited amount of data and can be applied in one-shot learning scenarios.

$k$th maximum-pooling can be defined as:

\begin{equation}
    P_{sort}(I_{i,j},k)= sort(R_{i,j})[k]
\end{equation}

where $sort$ is a simple sorting algorithm (sorting the elements in descending order) and $k$ is a parameter of the pooling method selecting the $k$th largest element in the region. If $k$ equals one \footnote{assuming indexing starts from one} the method results the original maximum pooling algorithm.

By this, one can ensure that a feature will be selected, which has multiple, large responses in a region. Features appearing locally, which might result overfitting on the data will be omitted.

\section{Results of  $k$th Maximum-Pooling}\label{SimResults}

To present that the algorithm can help in the training of convolutional networks we executed simulations using a simple and ocmmonly cited dataset datasets: MNIST \cite{lecun1998mnist} with a three layered convolutional network (containing 8,32,64 $3\times3$ convolution kernels),
We also have to note that there are many other methods to increase the generalization power of a network, like batch-normalization\cite{ioffe2015batch} and dropout\cite{srivastava2014dropout} and SeLUs\cite{klambauer2017self} all the results presented here were trained with networks where all three of these elements were present.

\begin{figure}[h!tb]
\centering{
\subfigure{\includegraphics[width=3.5in]{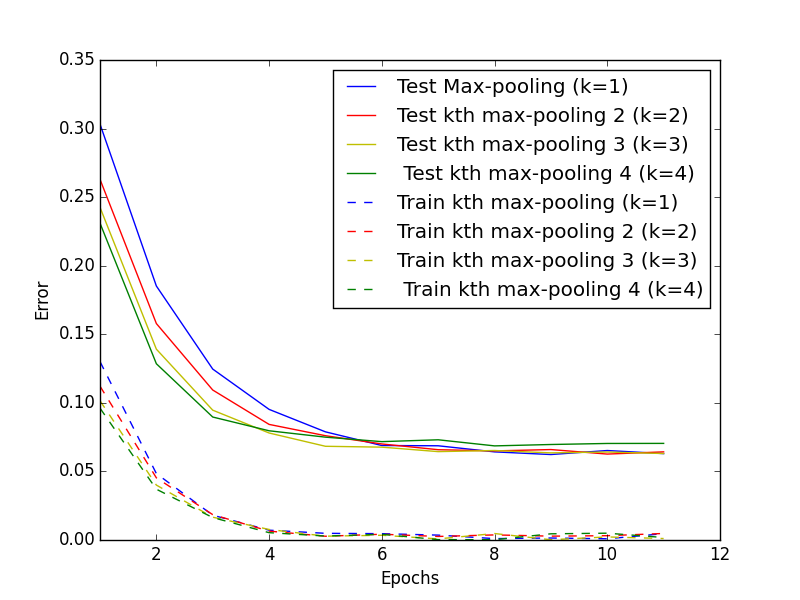}}\\
\caption{The plot shows the error rate on the independent test set on the MNIST dataset. Test error are displayes by normal, and train errors by a dashed line. As it can be seen with the increase of the $k$ parameter the network converges faster.}\label{MNISTResults}
}
\end{figure}

We have created a simple network containing three convolutional layers (with 8,32 and 64 features in the layers) followed by a fully connected layer. Each layer contained convolutions with $3\times3$ kernels and stride of one and pooling layers with $3\times3$ kernels and a stride of two. We have used maximum pooling ($k=1$) and  $k$th maximum-pooling with $k$ equals $2$, $3$ and $4$.
The results of the network, the error on the independent test set can be seen on Fig. \ref{MNISTResults}. These results show the averaged error on the independent test set on the MNIST dataset averaged from $50$ test runs. 
A summary of the numerical values can be found in Table \ref{tab:results}.
Now difference was seen in case of the train accuracy.
As it can be seen the network converges much faster compared to maximum pooling, which means better generalization, since train accuracies were similar, as not the maximum activation, but the second, third or fourth highest activation is selected. It is also worthwhile to note that if $k$ is fairly large, the convergence of the network will become fast, but the final error on the test set (and also on the training set) remains higher. For higher $k$ values than four the performance of the network decreased drastically (below 40\% test accuracy and because this the results are not displayed). After a given point the network can learn fairly general features but cannot reach high accuracy, because the features are too general and not specific enough.

\begin{table}[h!]
  \centering
  
  \begin{tabular}{|l||c|c|c|c|}
  \hline
     & k=1& k=2  & k=3 &k=4 \\
    \hline
    \makecell{MNIST \\ 1 epoch} & 30.2\% & 26.3\%  & 24.6\% & \textbf{22.8\%} \\
    \hline
    \makecell{MNIST \\ 3 epochs} & 12.4\% & 10.5\% & 9.5\% & \textbf{9.3\%} \\
     \hline
    \makecell{MNIST \\ 10 epochs} & \textbf{6.1}\% & 6.3\% & 6.3\% &  7.0\% \\
     \hline
  \end{tabular}
  \caption{Results achieved on the MNIST dataset. As it can be seen the application of  $k$th maximum-pooling can increase the converge speed of the network but results slightly lower accuracy in case of large dataset. Until only a smaller number of input data was showed to the network (1 and 3 epoch)  $k$th maximum-pooling performed better}.
  \label{tab:results}
\end{table}

\section{Sorted Pooling}

As it could be seen, a balance between the selection of general and specific kernels is important and can result faster convergence and higher accuracy. To ensure that the network has the ability to create this balance, we introduce sorted pooling. Sorted pooling can be seen as the generalization of  $k$th maximum-pooling introduced in Section \ref{SortedPooling}, where we select not only the $k$ largest element, but the top $K$ elements and create a weighted summation of them:
\begin{equation}\label{weightedpooling}
    P_{weight}(I_{i,j})=  \sum_{k=1}^{K} W_k P_{sort}(R_{i,j},k) 
\end{equation}

Where $W_k$ are weights of the activation functions, which ca be learned by the network. We have to ensure that $W_k\geq0$ for every $k$ and $\sum W_k = 1$. This can easily be ensured by applying a softmax function on the weights:

\begin{equation}
w_k=\frac{ e^{w^*_k} }{ \sum_{l=1}^{K} e^{w^*_l}  }
\end{equation}
where $W_k$ are the normalized weights used in Eq. \ref{weightedpooling} and $w^*_k$ are the original weights learned by the network through gradient descent methods. The initialization of these variables can be done by using teh same value for each $W_k$ or by an exponentially decaying distribution using smaller and smaller weights for more and more general features, (larger $k$ parameters).

The parameters $w^*_k$ are initialized and learned independently for every pooling operation in each layer and every feature (depth on the input data). But the same parameters are applied along the width and height of the input images.

Sorted pooling can also be seen as a general mathematical operation connecting convolution and pooling. In case of $K=1$ the method results maximum pooling, meanwhile if $K$ is the same as the number of elements in the pooling region, the result is similar to a convolution, where the activation will be the weighted summation of all the input elements, but the input elements are in decreasing order according to their responses instead of representing local information. If $K$ equals the size of the pooling region and the weights are one over the size of the pooling region we will get average pooling back. From this one can also see, how the parameter $K$ can create the balance between sparse and specific update and general regional update of the weights.

\section{Results with Sorted Pooling}\label{CombinedPooling}

\begin{figure}[h!tb]
\centering{
\subfigure{\includegraphics[width=3.5in]{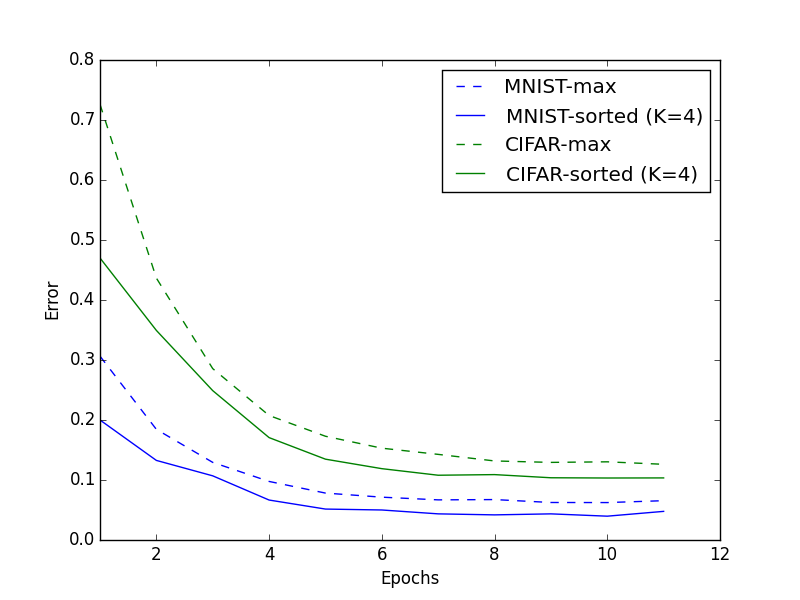}}\\
\caption{The comparison of maximum and sorted pooling on the MNIST and CIFAR datasets. As it can be seen from the results sorted pooling results lower accuracy on both datasets.
}\label{CIFArResults}
}
\end{figure}

We have trained a modified version of the VGG-16\cite{simonyan2014very} architecture on the CIFAR-10\cite{krizhevsky2014cifar} dataset in which the number of kernels and convolution sizes remained the same, but pooling operations were changed to $3\times3$ kernels with stride of 2.
The independent test error for the MNIST dataset with the architecture described in Section \ref{SimResults} and on the CIFAR-10 dataset with the VGG-16 architecture are plotted and compared for the maximum and sorted pooling methods on Figure \ref{CIFArResults}.
To be fair we have to note that we had to resize the input images to be compatible with the architecture, but this is a common step in networks used for classification, where small objects in the background of the scene had to be resized for classification to ensure compatibility with the network structure.

\begin{table}[h!]
  \centering
  \begin{tabular}{|l||c|c|}
  \hline
     & Max pooling & Sorted pooling (K=4) \\
    \hline
    \makecell{MNIST \\ 1 epoch} & 30.2\% & \textbf{22.5\%}  \\
    \hline
    \makecell{MNIST \\ 3 epochs} & 12.4\% &  \textbf{8.1\%} \\
     \hline
    \makecell{MNIST \\ 10 epochs} & 8.1\% & \textbf{5.8\%} \\
    \hline \hline
    \makecell{CIFAR \\ 1 epoch} & 67.3\% &  \textbf{49.2\%} \\
    \hline
    \makecell{CIFAR \\ 3 epochs} & 28.4\% &  \textbf{20.8\%} \\
     \hline
    \makecell{CIFAR \\ 10 epochs} & 18.5\%  & \textbf{14.8\%} \\
    \hline
  \end{tabular}
  \caption{Results achieved on different dataset and different architectures. As it can be seen sorted pooling algorithm performs slightly better in all cases and decreases training time with all architectures. }
  \label{tab:results2}
\end{table}

\begin{figure}[h!tb]
\centering{
\subfigure{\includegraphics[width=3.5in]{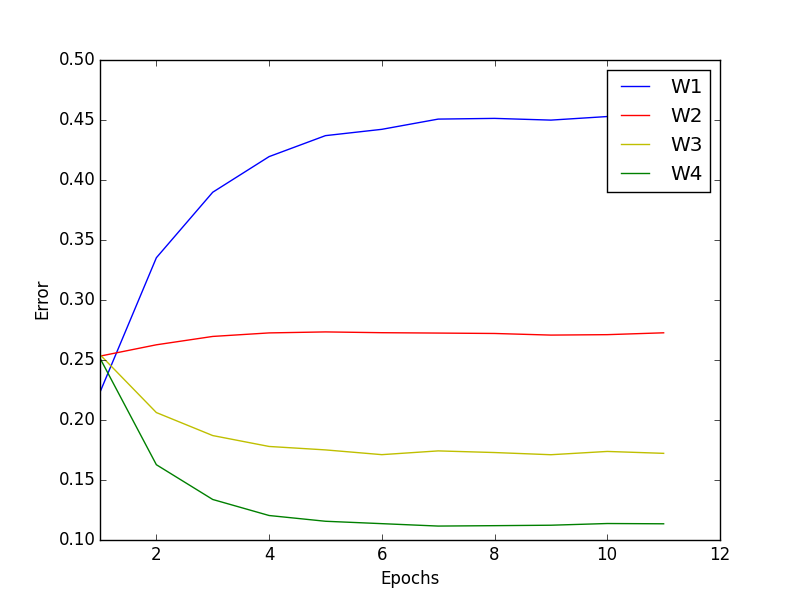}}\\
\caption{The evolution of the average of the weights for different $k$ values on the CIFAR-10 dataset are depicted on this figure. The weights are set to the same value at the beginning, and the network learn during training how to weight the largest and the smaller responses in each kernel. These values are average values over the whole architecture, the weights of each kernel in each layer was averaged, which hides the possible difference of certain kernels, but inevitable shows the the network uses information for the second and third largest responses as well.
}\label{wdist}
}
\end{figure}

As it can be seen from the results the network can indeed learn the weighting of the top $K$ elements in a region. One could expect that the weights will converge to having a value of $1$ at $k=1$ and zero at other indices, meaning that the algorithm converges to max pooling. But this is not the case and other indices also had non zero weights, although the values were decreasing with the increase of $k$. The distribution of average weights for different $k$ values on the CIFAR-10 dataset can be seen on Fig. \ref{wdist}.
We also have to not that this problem does not solve the problem of overfitting completely, and test accuracy can decrease after a number of steps (along with increasing train accuracy), but sorted pooling definitely helps and the results are better as in case of maximum pooling.

\subsection{Results in one-shot learning scenarios}

\begin{figure}[h!tb]
\centering{
\subfigure{\includegraphics[width=3.5in]{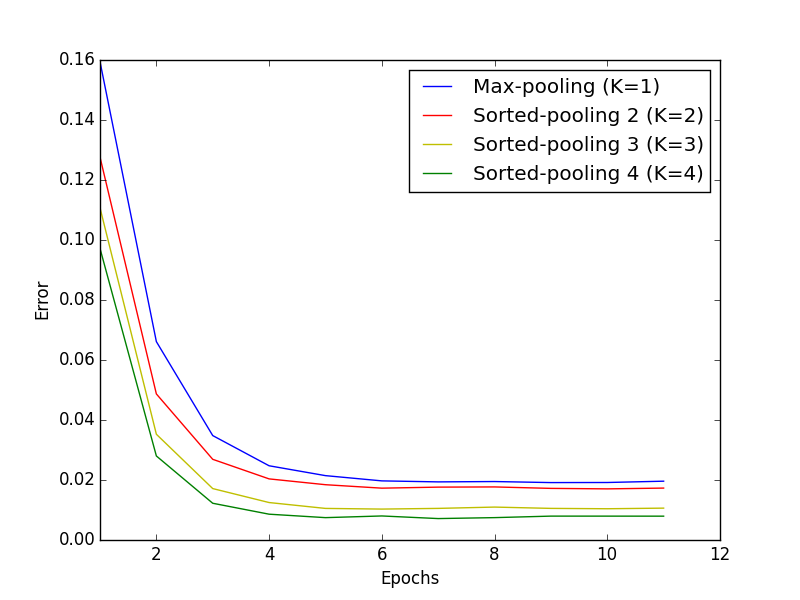}}\\
\caption{The independent test error on the Omniglot dataset for one-shot learning scenarios with matching-networks. The parameters of the simulation were the same as introduced in \cite{vinyals2016matching}.
}\label{OneShotResults}
}
\end{figure}

Feature generalization might be especially important in cases where large amount of data is not available. 
The extreme case of limited data are one-shot learning scenarios, where only one or a few instances of data is available, but there are also other problems in practice when data collection is cumbersome or expensive. Because of this we have selected an architecture which performs well in one-shot learning scenarios: matching networks\cite{vinyals2016matching} and investigated how sorted pooling affects the accuracy of such architectures. The test accuracy on the Omniglot dataset  \cite{lake2015human} calculating one-shot, five-way accuracy can be seen on Fig. \ref{OneShotResults} and in Table \ref{tab:results}. The architecture, train and test setups were the same as in \cite{vinyals2016matching} apart from the extension of the pooling kernels to $3\times3$ regions.

As it can be seen the increase of the $K$ parameter helped the network to find general features and it resulted an overall increase of accuracy. Compared to the MNIST and CIFAR scenarios where large amount of data could balance out the overfitted kernels, in case of one-shot learning sorted pooling resulted a higher increase in overall accuracy. We also have to note that the test accuracy is higher than the accuracy of the original approach \cite{vinyals2016matching}, which resulted  $98.1\%$ accuracy and in this case the accuracy was $98.9\%$.



\section{Conclusion}

In this paper we have introduced  $k$th maximum-pooling and sorted pooling which can be considered as an extension of the classical maximum pooling method and can also defined as a connecting link between maximum pooling, average pooling and convolution.
As it can be seen this method can help the network to find general features which is beneficial during training to increase the generalization power of the architecture. Sorted pooling resulted higher accuracy on all datasets with all architectures than maximum pooling. 

$k$th maximum-pooling and sorted pooling can be especially important in scenarios where only a limited amount of data is available. In this tasks both  $k$th maximum and sorted pooling approaches perform significantly better than traditionally applied pooling methods, in one-shot learning scenarios sorted pooling could further increase the accuracy of state of the art approaches and architectures, like matching networks.

\bibliographystyle{IEEEtran}
\bibliography{main}

\begin{thebibliography}{10}
\providecommand{\url}[1]{#1}
\csname url@samestyle\endcsname
\providecommand{\newblock}{\relax}
\providecommand{\bibinfo}[2]{#2}
\providecommand{\BIBentrySTDinterwordspacing}{\spaceskip=0pt\relax}
\providecommand{\BIBentryALTinterwordstretchfactor}{4}
\providecommand{\BIBentryALTinterwordspacing}{\spaceskip=\fontdimen2\font plus
\BIBentryALTinterwordstretchfactor\fontdimen3\font minus
  \fontdimen4\font\relax}
\providecommand{\BIBforeignlanguage}[2]{{%
\expandafter\ifx\csname l@#1\endcsname\relax
\typeout{** WARNING: IEEEtran.bst: No hyphenation pattern has been}%
\typeout{** loaded for the language `#1'. Using the pattern for}%
\typeout{** the default language instead.}%
\else
\language=\csname l@#1\endcsname
\fi
#2}}
\providecommand{\BIBdecl}{\relax}
\BIBdecl

\bibitem{krizhevsky2012imagenet}
A.~Krizhevsky, I.~Sutskever, and G.~E. Hinton, ``Imagenet classification with
  deep convolutional neural networks,'' in \emph{Advances in neural information
  processing systems}, 2012, pp. 1097--1105.

\bibitem{vinyals2016matching}
O.~Vinyals, C.~Blundell, T.~Lillicrap, D.~Wierstra \emph{et~al.}, ``Matching
  networks for one shot learning,'' in \emph{Advances in Neural Information
  Processing Systems}, 2016, pp. 3630--3638.

\bibitem{santoro2016one}
A.~Santoro, S.~Bartunov, M.~Botvinick, D.~Wierstra, and T.~Lillicrap,
  ``One-shot learning with memory-augmented neural networks,'' \emph{arXiv
  preprint arXiv:1605.06065}, 2016.

\bibitem{geman2008neural}
S.~Geman, E.~Bienenstock, and R.~Doursat, ``Neural networks and the
  bias/variance dilemma,'' \emph{Neural Networks}, vol.~4, no.~1, 2008.

\bibitem{ioffe2015batch}
S.~Ioffe and C.~Szegedy, ``Batch normalization: Accelerating deep network
  training by reducing internal covariate shift,'' in \emph{International
  Conference on Machine Learning}, 2015, pp. 448--456.

\bibitem{srivastava2014dropout}
N.~Srivastava, G.~E. Hinton, A.~Krizhevsky, I.~Sutskever, and R.~Salakhutdinov,
  ``Dropout: a simple way to prevent neural networks from overfitting.''
  \emph{Journal of machine learning research}, vol.~15, no.~1, pp. 1929--1958,
  2014.

\bibitem{he2016deep}
K.~He, X.~Zhang, S.~Ren, and J.~Sun, ``Deep residual learning for image
  recognition,'' in \emph{Proceedings of the IEEE conference on computer vision
  and pattern recognition}, 2016, pp. 770--778.

\bibitem{springenberg2014striving}
J.~T. Springenberg, A.~Dosovitskiy, T.~Brox, and M.~Riedmiller, ``Striving for
  simplicity: The all convolutional net,'' \emph{arXiv preprint
  arXiv:1412.6806}, 2014.

\bibitem{kingma2013auto}
D.~P. Kingma and M.~Welling, ``Auto-encoding variational bayes,'' \emph{arXiv
  preprint arXiv:1312.6114}, 2013.

\bibitem{goodfellow2014generative}
I.~Goodfellow, J.~Pouget-Abadie, M.~Mirza, B.~Xu, D.~Warde-Farley, S.~Ozair,
  A.~Courville, and Y.~Bengio, ``Generative adversarial nets,'' in
  \emph{Advances in neural information processing systems}, 2014, pp.
  2672--2680.

\bibitem{fei2006one}
L.~Fei-Fei, R.~Fergus, and P.~Perona, ``One-shot learning of object
  categories,'' \emph{IEEE transactions on pattern analysis and machine
  intelligence}, vol.~28, no.~4, pp. 594--611, 2006.

\bibitem{ren2015faster}
S.~Ren, K.~He, R.~Girshick, and J.~Sun, ``Faster r-cnn: Towards real-time
  object detection with region proposal networks,'' in \emph{Advances in neural
  information processing systems}, 2015, pp. 91--99.

\bibitem{he2017mask}
K.~He, G.~Gkioxari, P.~Doll{\'a}r, and R.~Girshick, ``Mask r-cnn,'' \emph{arXiv
  preprint arXiv:1703.06870}, 2017.

\bibitem{murray2014generalized}
N.~Murray and F.~Perronnin, ``Generalized max pooling,'' in \emph{Proceedings
  of the IEEE Conference on Computer Vision and Pattern Recognition}, 2014, pp.
  2473--2480.

\bibitem{lee2016generalizing}
C.-Y. Lee, P.~W. Gallagher, and Z.~Tu, ``Generalizing pooling functions in
  convolutional neural networks: Mixed, gated, and tree,'' in \emph{Artificial
  Intelligence and Statistics}, 2016, pp. 464--472.

\bibitem{lecun1995convolutional}
Y.~LeCun, Y.~Bengio \emph{et~al.}, ``Convolutional networks for images, speech,
  and time series,'' \emph{The handbook of brain theory and neural networks},
  vol. 3361, no.~10, p. 1995, 1995.

\bibitem{hou2007saliency}
X.~Hou and L.~Zhang, ``Saliency detection: A spectral residual approach,'' in
  \emph{Computer Vision and Pattern Recognition, 2007. CVPR'07. IEEE Conference
  on}.\hskip 1em plus 0.5em minus 0.4em\relax IEEE, 2007, pp. 1--8.

\bibitem{simonyan2014very}
K.~Simonyan and A.~Zisserman, ``Very deep convolutional networks for
  large-scale image recognition,'' \emph{arXiv preprint arXiv:1409.1556}, 2014.

\bibitem{lecun1998mnist}
Y.~LeCun, ``The mnist database of handwritten digits,'' \emph{http://yann.
  lecun. com/exdb/mnist/}, 1998.

\bibitem{klambauer2017self}
G.~Klambauer, T.~Unterthiner, A.~Mayr, and S.~Hochreiter, ``Self-normalizing
  neural networks,'' \emph{arXiv preprint arXiv:1706.02515}, 2017.

\bibitem{krizhevsky2014cifar}
A.~Krizhevsky, V.~Nair, and G.~Hinton, ``The cifar-10 dataset,'' \emph{online:
  http://www. cs. toronto. edu/kriz/cifar. html}, 2014.

\bibitem{lake2015human}
B.~M. Lake, R.~Salakhutdinov, and J.~B. Tenenbaum, ``Human-level concept
  learning through probabilistic program induction,'' \emph{Science}, vol. 350,
  no. 6266, pp. 1332--1338, 2015.

\end{thebibliography}
\end{document}